\pdfoutput=1
\documentclass[letterpaper]{article}
\PassOptionsToPackage{hyphens}{url}
\usepackage{aaai2027}
\usepackage{url}
\usepackage{graphicx}
\usepackage{natbib}
\usepackage{caption}
\usepackage{amsmath}
\usepackage{amssymb}
\usepackage{multirow}
\usepackage{xspace}
\usepackage{booktabs}
\usepackage{tabularx}
\usepackage{array}
\usepackage{xcolor}
\usepackage{tikz}

\title{PowerFix: Physics-Guided Iterative Repair for LLM-Assisted Power Converter Design}
\author{
    Can Wang\equalcontrib\textsuperscript{\rm 1},
    Shi Qiu\equalcontrib\textsuperscript{\rm 1,2},
    Fanfan Lin\textsuperscript{\rm 3}
}
\affiliations{
    \textsuperscript{\rm 1}Xi'an Jiaotong-Liverpool University, Suzhou, China\\
    \textsuperscript{\rm 2}New York University, New York, USA\\
    \textsuperscript{\rm 3}ZJU-UIUC Institute, Zhejiang University, Haining, China\\
    can.wang22@student.xjtlu.edu.cn, sq2335@nyu.edu, fanfanlin@intl.zju.edu.cn
}

\nocopyright

\begin{document}
\flushbottom
\maketitle

\begin{abstract}
Large language model (LLM) agents are increasingly being explored for automated engineering design. However, repairing power-converter designs that fail engineering validation remains challenging: the system must identify physical causes, locate responsible design variables, and resolve violations without compromising other constraints. We introduce PowerFix, a physics-guided iterative repair framework that converts validation failures into structured diagnoses and executable patches while preserving topology and design requirements. PowerDiagGraph links failed checks to possible physical causes and constrains repair routes. An LLM prioritizes these routes, while a deterministic compiler translates selected repairs into authorized field-level edits. Candidates are validated on isolated design copies and committed only after all required engineering checks pass; failed validation can guide replanning. On our constructed 96-task flyback repair benchmark, comprising injected single faults, coupled faults, and held-out natural failures from our design generator, PowerFix achieves 78.7\% verified repair success rate (VRSR), compared with 30.1\% for the strongest of three LLM baselines. Three additional replanning rounds increase VRSR to 88.4\%. These offline results provide pilot-stage evidence within a simulation-based design workflow, with hardware prototype validation and transfer to a second converter topology as the next steps toward deployment.
\end{abstract}
\section{Introduction}
\label{sec:introduction}

Power converters connect energy sources and storage devices to electrical grids, regulating voltage, current, and power flow in renewable energy systems, electric transportation, and industrial automation \citep{bose2009power}. Their failures are a major reliability concern. In 2023, Toyota recalled 43,442 plug-in hybrid electric vehicles in the United States due to a DC--DC converter defect that could cause overheating (the recall report was revised in 2024) \citep{nhtsa2024toyota}. Beyond such safety risks, uncorrected engineering errors may cause severe hardware damage once a design is implemented. For example, an incorrect ripple-current calculation may result in the selection of a capacitor with an insufficient ripple-current rating. The converter may continue to operate normally at first, while excessive ripple current increases the capacitor temperature and accelerates its degradation \citep{wang2014capacitors}.

Recent advances in large language models (LLMs) have driven growing interest in LLM-assisted power electronics design. PE-GPT explored LLM-based converter design \citep{lin2024pegpt,lin2025pegpt}, while Huang et al.\ proposed a multi-agent framework for controller tuning and calibration \citep{huang2025multiagent}. LLM-based workflows nevertheless remain error-prone on engineering design tasks. On 269 switched-mode power-supply tasks, standalone GPT-4o and o3 failed on 85.6\% and 74.6\% of the cases, respectively; retrieval and iterative SPICE feedback increased o3's solve rate to 90.7\%, demonstrating the value of engineering evidence \citep{nau2026smps}.

However, feedback-driven refinement does not by itself provide a structured repair process for a design that has already failed engineering validation. A repair system must connect observed violations to their underlying physical causes, identify the responsible design variables, restrict modifications to admissible fields, and verify that correcting the reported failure does not leave or introduce violations elsewhere. Existing workflows do not explicitly combine these requirements into a validation-gated repair procedure.

\begin{figure*}[t!]
    \centering
    \includegraphics[width=0.88\textwidth]{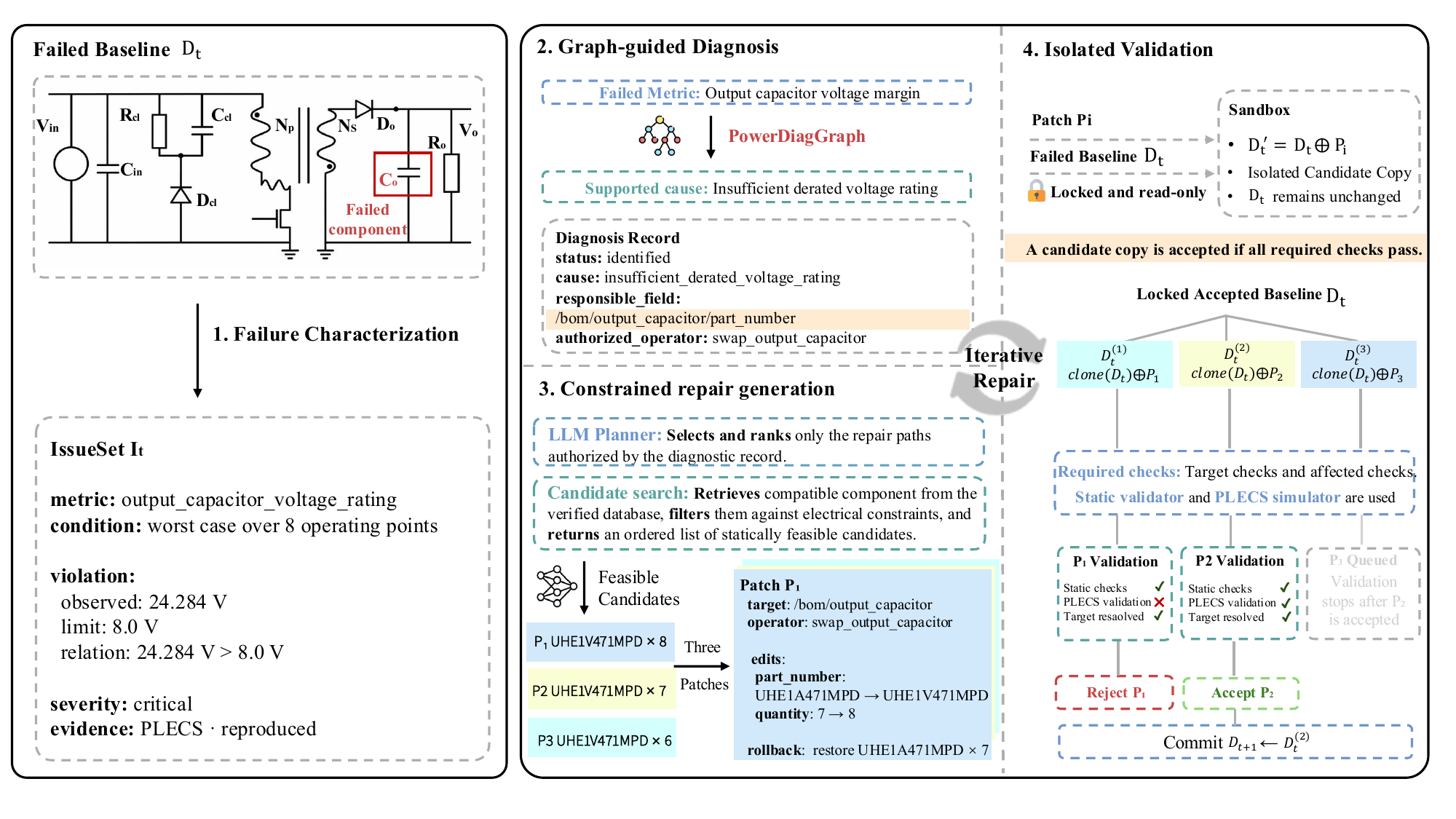}
    \caption{
    Overview of \textbf{PowerFix}. PowerFix performs iterative repair through failure characterization, graph-guided diagnosis, constrained repair generation, and isolated validation.
    }
    \label{fig:powerfix_overview}
\end{figure*}

To address this problem, we introduce \textbf{PowerFix}, a physics-guided iterative repair framework for LLM-assisted power converter design. Its four-stage workflow consists of Failure Characterization, Graph-Guided Diagnosis, Constrained Repair Generation, and Isolated Validation.

The framework follows three principles.

\noindent\textbf{1. Physics-grounded diagnosis.}
PowerFix converts validation failures into structured evidence and traces them through a topology-specific diagnostic graph to supported physical causes and admissible repair routes.

\noindent\textbf{2. Constrained executable repair.}
An LLM prioritizes repair routes authorized by the diagnosis, while deterministic repair operators compile the selected routes into field-level edits. The LLM therefore does not directly perform unrestricted modification of the converter state.

\noindent\textbf{3. Validation-before-commit.}
Each candidate patch is evaluated on an isolated copy of the current design and committed only if all required static, analytical, and circuit-simulation checks pass. Otherwise, the baseline remains unchanged and the validation evidence can guide a subsequent repair attempt.

Our contributions are threefold:
\begin{itemize}
    \item We study how to repair failed converter designs so that they pass every required engineering check, and show why fixing the reported fault alone is insufficient.
    \item We introduce PowerFix, which links physics-guided diagnosis to authorized repair routes, combines LLM planning with deterministic patch compilation, and accepts patches only after they pass validation on isolated design copies. 
    \item We build a 96-task flyback benchmark of single, coupled, and held-out natural failures, on which PowerFix reaches 78.7\% group-weighted VRSR against 30.1\% for the strongest of three LLM baselines (88.4\% with three replanning rounds). Ablations test each stage; planning studies compare success and cost.

\end{itemize}
\section{Related Work}
\label{sec:related}

\paragraph{Circuit and converter design.}
Converter optimization searches for trade-offs in efficiency, power
density, and cost \citep{burkart2017comparative}; repair starts from an
existing design that fails required checks.
LLM-based assistance spans EDA scripts and circuit generation
\citep{liu2023chipnemo,lai2025analogcoder}; \citet{pan2025survey} survey the area.
Power-electronics assistants support parameter and modulation design
\citep{liu2024chatgpt,lin2024pegpt,lin2025pegpt}. Others gather requirements
or tune controllers \citep{liu2026phia,huang2025multiagent}.
Simulation-guided refinement \citep{dong2026flyback,liu2026accessible}
includes direct netlist edits \citep{nau2026smps}; in PowerFix, the LLM
selects routes allowed by the diagnosis and a compiler writes the edits.

\paragraph{Fault diagnosis.}
\citet{gao2015diagnosis} review real-time fault diagnosis using models and
signals. PowerDiagGraph addresses failed design checks before hardware
is built: it links analytical and simulation evidence to physical causes,
which determine the permitted repair routes.

\paragraph{LLM feedback and program repair.}
Software patches can pass repair tests yet fail independent tests
\citep{smith2015overfitting}. RIRR captures a related gap: target fixes that
still fail other engineering checks.
Our Self-Refine baseline adapts iterative revision \citep{madaan2023selfrefine}
to engineering feedback; the Tool Agent uses actions and feedback, as in
ReAct \citep{yao2023react}. Reflexion uses reflections to guide retries
\citep{shinn2023reflexion}; PowerFix uses validation feedback to replan
within the existing diagnosis.
SWE-bench, SWE-agent, and RepairAgent study repository repair
\citep{jimenez2024swebench,yang2024sweagent,bouzenia2025repairagent}.
A software repair loop alone does not specify which physical changes
are permitted or which checks those changes require.

\section{The Repair Problem}
\label{sec:problem}

Repair starts from a design that fails the full engineering regression.
At step $t$, the failure set $\mathcal{I}_t$ (\texttt{IssueSet} in
Figure~\ref{fig:powerfix_overview}) records each failed metric,
its operating condition, observed and required values, and supporting
evidence. Diagnosis and validation treat the failures as one set,
because a fix for one can create another.

A patch $P$ modifies only authorized design fields. Its required checks
cover the reported failures and every check the modified fields can influence:
\begin{equation}
    \mathcal{C}_{\mathrm{req}}(P,\mathcal{I}_t)
    =
    \mathcal{C}_{\mathrm{target}}(\mathcal{I}_t)
    \cup
    \mathcal{C}_{\mathrm{affected}}(P) .
    \label{eq:creq}
\end{equation}
The second set catches other failures that the patch may leave unresolved
or introduce. Only a patch passing these checks can replace the current
design. Appendix~\ref{app:formalism} gives the design state, patch representation,
and full acceptance rule.

\section{PowerFix}
\label{sec:method}

Figure~\ref{fig:powerfix_overview} illustrates the four-stage repair loop.

\paragraph{Failure characterization.}
Patches may edit only primary fields, such as switching frequency, transformer winding parameters, and component selections. All other fields, including derived quantities, validation records, requirements, and the topology definition, are protected. Engineering tools recompute derived quantities. Failure characterization re-checks each reported violation against the current state and records it in the failure set $\mathcal{I}_t$.

\paragraph{Graph-guided diagnosis.}
PowerDiagGraph constrains the search for physical causes.
Its nodes represent failed metrics, intermediate physical quantities,
and possible causes. Arcs come from converter equations and design rules:
they show how a metric decomposes into quantities or how a cause produces
an effect. The flyback PowerDiagGraph
contains 174 nodes and 301 arcs. It is authored offline once per topology
and reused across tasks; its construction cost is excluded from the
reported per-task costs. PowerDiagGraph is fixed during inference and cannot
be extended by the model.

For each failed metric, diagnosis follows the corresponding subgraph to
possible causes and the quantities needed to evaluate them. It retrieves
design values and waveform features, then uses the topology's analytical
and simulation tools to compare measurements with physical predictions.
Each cause is marked \emph{supported}, \emph{contradicted}, or
\emph{unresolved}. When the evidence cannot distinguish competing causes,
PowerFix runs a targeted diagnostic simulation. The resulting record
contains the causes, evidence, responsible primary fields, and allowed
repair operations for this topology.
Appendix~\ref{app:method} (``A worked diagnosis'') gives a
MOSFET voltage-margin example in which simulation distinguishes
insufficient clamp absorption from excessive leakage energy.

\paragraph{Constrained repair generation.}
A repair dimension is an aspect of the design,
such as MOSFET selection. An operation specifies a permitted edit, such
as swapping that MOSFET. An option is a concrete choice within a dimension,
such as a catalog MOSFET. A repair route is one way of fixing the failure that the diagnosis
supports; candidate patches instantiate it with particular parts or parameters.

An atomic patch records one permitted operation, its arguments, and
the resulting field edits. Each edit names the target field and its current and proposed
values. The planning LLM chooses which allowed repair routes to try
first.
A deterministic compiler produces the field
edits and retrieves device properties from the component database when a
part is replaced. Several atomic patches can be combined into a joint
candidate for validation.

A hand-authored field-to-check map, separate from PowerDiagGraph, defines authorized fields and their associated checks. The validator takes the union of checks associated with the fields written by a patch and rejects any patch that writes an unlisted field.

In our experiments, all 29 authorized fields map to the full engineering regression over eight operating points, so every candidate uses the same
$\mathcal{C}_{\mathrm{req}}$. Equation~\ref{eq:creq} permits patch-specific
check selection, but we do not evaluate its potential savings. Appendix~\ref{app:method} describes an unused finer map with a different
field schema.

\paragraph{Isolated validation.}
Each candidate is applied to a copy of the current design, leaving the baseline
unchanged during validation. Checks run in order of increasing cost:
static checks on schema, field permissions, units, preconditions, and
component constraints; analytical checks on rating, duty-cycle, magnetic,
and thermal constraints; and circuit simulation over the required
operating conditions. The validation record includes the status of the
target failures, check outcomes, newly introduced violations, and measured
values. A candidate is committed only after every required check passes;
otherwise it is rejected and the baseline remains unchanged.

The system retains rejected candidates and their validation evidence.
When replanning is enabled, the planner reads this evidence and can choose
another repair route within the existing diagnosis. Otherwise, the search continues
through the preplanned candidates. The loop ends when a candidate is
accepted or the repair budget is exhausted.

\section{Evaluation Setup}
\label{sec:evaluation}

We evaluate PowerFix on 96 failed flyback designs. The single-fault
tasks cover every combination of three input specifications
(85--132, 180--265, and 85--265\,Vac), four output specifications spanning
10--24\,W, and six fault classes ($3\times4\times6=72$).
The fault classes are turns ratio, magnetizing inductance, magnetic
realization, MOSFET, secondary rectifier, and output capacitor.
Each task has a known violation and a recorded ground-truth repair.
Twelve coupled tasks introduce interacting fault pairs, where the first
visible violation does not identify the design error. The remaining
twelve are \emph{natural failures} from a design generator modeled on
PE-GPT \citep{lin2024pegpt}. Its LLM roles propose the system, component,
and magnetics designs; a fourth role integrates them, and engineering
tools validate the result. We retain the generated design values, source
outputs, and validation records. These tasks contain no injected faults
and have no unique correct repair. We did not use them when developing
or tuning PowerFix; they are scored on VRSR like the other two groups.
Appendix~\ref{app:benchmark} gives the construction.

\begin{table}[!b]
\centering
\small
\begin{tabular*}{\columnwidth}{@{\extracolsep{\fill}}lcccc@{}}
\toprule
Study & $R$ & \shortstack{Candidates\\/task} &
\shortstack{Options\\/dimension} & VRSR \\
\midrule
Primary & 0 & $\leq6$ & $\leq48$ & 78.7\% \\
Ablation & 0 & $\leq6$ & $\leq8$ & 72.7\% \\
Backbone & 5 & $\leq6$ & $\leq8$ & 75.9\% \\
\bottomrule
\end{tabular*}
\caption{PowerFix protocols on 96 tasks; VRSR uses GPT-5 Mini.
Primary options apply to joint repairs (single-fault repair: up to three
healthy candidates per route). Backbone: one candidate per round.}
\label{tab:protocols-main}
\end{table}

We compare complete regeneration, revision from validation feedback,
and revision through direct tool use (Appendix~\ref{app:baselines}).
\emph{Open-loop Regeneration} produces a complete replacement design.
Our \emph{Natural-language Self-Refine} baseline adapts iterative refinement
\citep{madaan2023selfrefine} to feedback from engineering checks; an LLM
parser turns its free-text edits into field updates.
The \emph{Unstructured Tool-Feedback Agent} also calls the same
component queries, analytical checks, and simulator as PowerFix, and
edits fields based on what those tools return. None of the three uses
PowerDiagGraph or constrained patches. All methods use the same tasks,
GPT-5 Mini-family backbone, component database, and validation conditions.
Table~\ref{tab:protocols-main} separates the primary comparison from the
component and backbone studies. Each of the latter compares variants
within its own protocol and full-system reference. Here $R$ counts
replanning rounds after the initial plan.
After each method finishes, a deterministic \emph{external scorer}
checks its final design against the full engineering regression, using
task records kept separately for evaluation. It is not an LLM judge,
and its results are not fed into the repair loop.

The methods have different execution limits and do not use matched
compute budgets. PowerFix makes about four times as many workflow
simulation calls as the Tool Agent (Table~\ref{tab:main}). The primary
comparison therefore measures complete workflows under their stated
limits. Component effects are tested separately with a common
candidate budget in the ablation.

Within each group, VRSR is the share of tasks whose final design resolves
the target failure and passes the full engineering regression.
The \emph{Target Issue Resolution Rate} (TIRR) measures target resolution
regardless of other checks. Both rates and per-task costs weight the
three group means equally; we call these \emph{group-weighted}.
We define the repair-induced regression rate as
$\mathrm{RIRR}=1-\mathrm{VRSR}/\mathrm{TIRR}$.
Error bars show 95\% bootstrap intervals; differences use paired resampling.
In the primary comparison, we resample base designs for controlled tasks
and individual tasks for natural failures.

\section{Results}
\label{sec:results}

\paragraph{Repair effectiveness.}
\label{sec:results-main}

\begin{figure}[!b]
    \centering
    \includegraphics[width=\columnwidth,trim=0bp 4bp 0bp 6.5bp,clip]{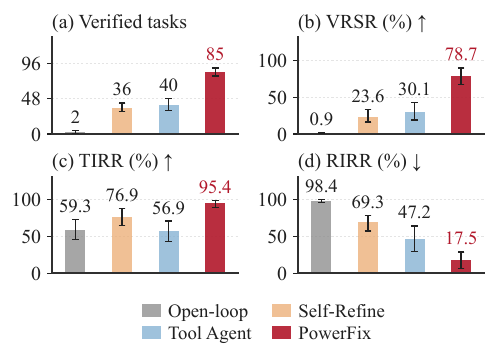}
    \caption{Primary results on 96 tasks: raw verified counts and
    group-weighted VRSR and TIRR. RIRR is the share of target fixes that
    fail other checks.}
    \label{fig:repair-main}
\end{figure}

\begin{figure}[!t]
    \centering
    \includegraphics[width=\columnwidth,trim=0bp 4bp 0bp 7.5bp,clip]{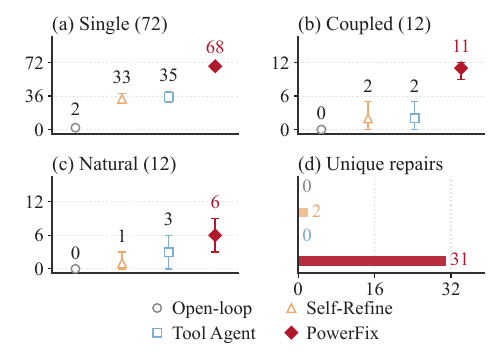}
    \caption{Verified tasks by group; parentheses give group sizes.
    Unique repairs are tasks verified by only one method.}
    \label{fig:group-repairs}
\end{figure}

PowerFix reaches 78.7\% VRSR against 30.1\% for the strongest baseline
(Figure~\ref{fig:repair-main}), a gain of 48.6 percentage points
(95\% CI $[33.8,63.0]$). It verifies 85 of 96 tasks (88.5\% raw);
group weighting gives equal importance to the two smaller groups.
Thirty-one tasks are repaired by PowerFix alone
(Figure~\ref{fig:group-repairs}(d)).

Self-Refine resolves more targets than the Tool Agent, but more of its
target fixes still fail other checks; their ranking reverses under VRSR
(Figure~\ref{fig:repair-main}(b--d)). The largest gap is on coupled faults:
11 of 12 repairs versus two (Figure~\ref{fig:group-repairs}(b)). On natural
failures, the paired interval for the gain is wide and includes zero
($[-8.3,58.3]$ percentage points), so we do not claim a significant gain.

\paragraph{Ablation.}
\label{sec:results-ablation}

\begin{figure}[!b]
    \centering
    \includegraphics[width=\columnwidth,trim=0bp 6bp 0bp 1bp,clip]{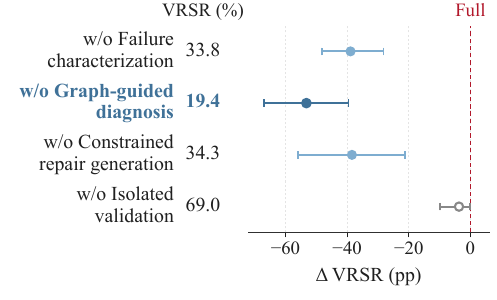}
    \caption{Ablation at $R=0$: VRSR differences from its own reference
    (72.7\%; red line; Table~\ref{tab:protocols-main}).}
    \label{fig:ablation}
\end{figure}

All four ablations share one full-system reference: 77 verified tasks
and 72.7\% VRSR (Figure~\ref{fig:ablation}). Under this protocol, the
planner selects one or two dimensions from the smaller option pool
in Table~\ref{tab:protocols-main}; the primary planner ranks all supplied
routes. The option limit, selection policy, and order of candidate
filtering differ between these studies, so their reference gap does not
measure the effect of the option limit alone (Appendix~\ref{app:protocols}).

Removing graph-guided diagnosis causes the largest drop (53.2 points),
with only one coupled repair and no natural repairs. Removing failure
characterization or constrained repair generation costs 38--39 points.
All three paired intervals exclude zero.

Removing isolated validation changes VRSR little but eliminates rejection:
the system hands back a design for all 96 tasks, and the external scorer
marks 22 invalid. The full system instead withholds 19 unrepaired outputs.
Without this stage, an engineer would receive the invalid designs as
completed repairs.

\paragraph{Repair-dimension selection.}
\label{sec:results-selection}

We compare the LLM planner with fixed-priority and random selection on
31 tasks that permit repairs in multiple dimensions. Each policy selects
two permitted dimensions per round. Five seeds give 155 trajectories per
policy (Appendix~\ref{app:top2}).

\begin{figure}[!htb]
    \centering
    \includegraphics[width=\columnwidth,trim=0bp 7.5bp 0bp 1bp,clip]{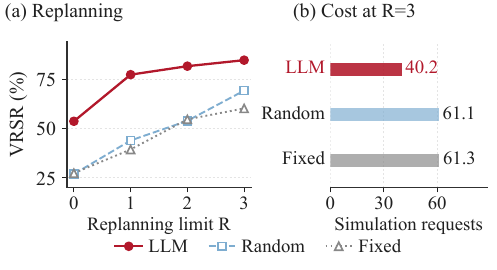}
    \caption{Repair selection (LLM: GPT-5 Mini): group-weighted
    (a) VRSR and (b) operating-point requests at $R=3$, including cache hits.}
    \label{fig:selection-main}
\end{figure}

\begin{figure}[!b]
    \centering
    \includegraphics[width=\columnwidth,trim=0bp 10bp 0bp 0.5bp,clip]{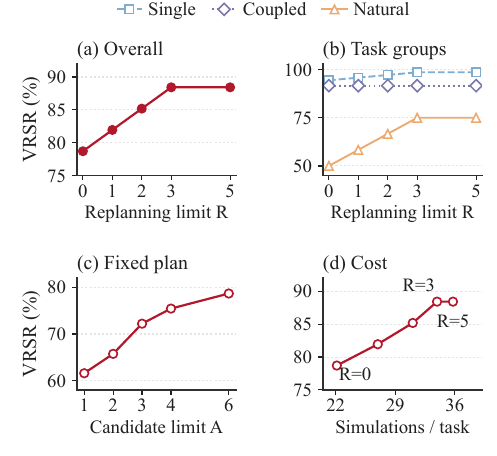}
    \caption{Budget effects on 96 tasks: (a) overall and (b) group
    repair rates with replanning; (c) candidate attempts at $R=0$;
    (d) replanning versus mean workflow simulations. Axis ranges differ.
}
    \label{fig:retry-main}
\end{figure}

The LLM planner has the highest VRSR at every reported budget and
requests about a third fewer simulations at $R=3$
(Figure~\ref{fig:selection-main}). Recorded active repair time is longer
than for the fixed policy (163 versus 125 seconds); it depends on cache
reuse and execution order.

\paragraph{Repair budget.}
\label{sec:results-robust}

On the full 96-task benchmark, we vary the preplanned candidate-attempt
limit $A$ and the additional replanning limit $R$
(Figure~\ref{fig:retry-main}). These budgets separate searching within
an initial plan from choosing another route after validation fails.

Three replanning rounds raise VRSR to 88.4\%, verifying 91 of 96 tasks
(Figure~\ref{fig:retry-main}(a)). Most of the gain comes from three
additional natural repairs: this 12-task group receives the same weight
as the 72 single-fault tasks (Figure~\ref{fig:retry-main}(b)). Two further
rounds add simulation work without another verified repair.
Appendix~\ref{app:budgets} lists counts and costs.

\paragraph{Backbone robustness.}
\label{sec:results-backbone-main}

The backbone study follows the protocol in Table~\ref{tab:protocols-main}:
$R=5$, one candidate per round, and up to eight options per dimension.
GPT-5 Mini provides its 75.9\% reference.

\begin{figure}[!htb]
    \centering
    \includegraphics[width=\columnwidth,trim=0bp 4bp 0bp 0.5bp,clip]{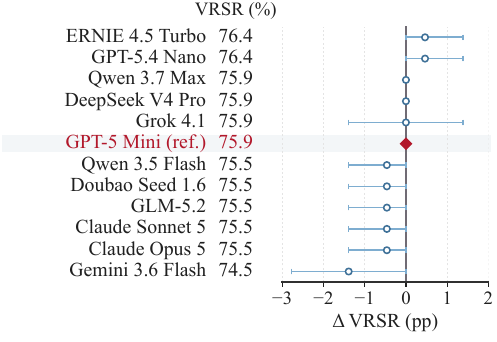}
    \caption{Backbone differences from its own GPT-5 Mini reference
    (75.9\% VRSR; $R=5$; Table~\ref{tab:protocols-main}).}
    \label{fig:backbone-main}
\end{figure}

Across twelve LLMs, VRSR spans 1.9 points
(Figure~\ref{fig:backbone-main}; Appendix~\ref{app:backbones}); the lower end of each paired 95\% interval stays
within five points of the reference. Scorer outcomes agree
on 91 of 96 tasks: only five single-fault tasks differ. All models leave
the same two coupled and five natural tasks unverified and request
about 18 to 19 operating-point evaluations per task.

\paragraph{Cost.}
\label{sec:results-cost}

PowerFix shifts work toward simulation: about 3\% of the Tool Agent's
tokens but four times its workflow simulations (Table~\ref{tab:main}).
At $R=0$, one planning call orders the routes; enumeration, patch
compilation, and validation then run deterministically. Mean
LLM calls rise to 1.54 at $R=3$ and 1.77 at $R=5$. Costs include failed
tasks but exclude initial graph and authorization-map authoring.
Deployment needs a setup budget for each topology and ongoing capacity
for updates and simulation.

Table~\ref{tab:main} counts logged simulation calls;
Figure~\ref{fig:selection-main} reports requested operating-point
evaluations, including cache hits (Appendix~\ref{app:protocols}).
These counts are not interchangeable.

\begin{table}[!htb]
\centering
\begin{tabular}{@{}lrrr@{}}
\toprule
Method & Simulations & LLM calls & Tokens (k) \\
\midrule
Open-loop   & 1.74 & 4.00 & 37.10 \\
Self-Refine & 2.84 & 5.23 & 33.44 \\
Tool Agent  & 5.31 & 11.08 & 82.86 \\
\textbf{PowerFix} & 22.13 & \textbf{1.00} & \textbf{2.73} \\
\bottomrule
\end{tabular}
\caption{Group-weighted primary cost per task ($R=0$ for PowerFix),
including failures. Simulation and LLM counts are workflow requests
(Appendix~\ref{app:token_audit}).}
\label{tab:main}
\end{table}

\paragraph{Remaining failures.}
\label{sec:results-failures}

Five tasks remain unverified at $R=3$: one single-fault, one coupled,
and three natural. At $R=0$, the one unverified coupled case combines
magnetizing-inductance and MOSFET faults in a 12\,V, 2\,A design at
85--132\,Vac. None of its three candidate bundles passes the full engineering
regression. One resolves the reported targets but still violates the
output-ripple limit at low line and full load. This case exposes a
limit of the available repair routes. Natural failures are also
difficult: they have no unique ground-truth repair and can admit several
plausible causes. Self-Refine's free-form edits verify three tasks missed by PowerFix
at $R=0$.

\section{Simulation-Stage Pilot and Deployment Path}
\label{sec:deployment}

\paragraph{Use in a design workflow.}
PowerFix runs after an LLM-generated design fails the full engineering
regression. It hands the engineer a verified patch or a rejection with supporting
evidence; the engineer retains sign-off authority. Each patch records the fields
written, their old and proposed values, the operation used, and the rollback
edits that restore the old values. Requirements and topology definitions remain
protected; the team approves the editable-field set before running repairs.

For the output-capacitor example in Figure~\ref{fig:powerfix_overview}, the
accepted record ($P_2$) replaces \texttt{UHE1A471MPD} with
\texttt{UHE1V471MPD} and explicitly retains quantity 7. This unchanged
quantity counts as a write in the profile below. The patch detail box in
Figure~\ref{fig:powerfix_overview} shows the rejected $P_1$, which uses eight capacitors
and fails simulation on a copy; the baseline stays unchanged. 

\paragraph{Patch size.}
Among 85 accepted submissions at $R=0$, counting each joint repair once,
41 explicitly write at most two distinct fields, 41 write 25--26, and
three write three. All large submissions include magnetics redesign,
which writes 25 core, winding, and primary-clamp fields; seven also
write a MOSFET or rectifier catalog reference. All writes are authorized,
and every submission includes rollback edits. Appendix~K details the
distribution, operation coverage, and counting rules. Review time and
engineer acceptance remain unmeasured.

\paragraph{Simulation-stage pilot.}
Engineers use simulation to refine converter designs and check operating
limits before building hardware. PowerFix automates repair and revalidation
within this stage. We evaluated it offline on the same 12 held-out failures from
design-generator runs: it repaired 6 (50\%) under the primary protocol
and 9 (75\%) with three replanning rounds. Each repaired design passed
the full engineering regression across eight operating points. These
results measure recovery within the simulation workflow; engineer
acceptance and hardware performance remain untested. We plan two next steps.

\emph{(i)~Hardware prototype validation.} We will build the repaired
24\,V, 1\,A flyback in Figure~\ref{fig:powerfix_overview} and measure peak device
voltage, output ripple, and efficiency across the same eight operating
points. We will compare measurements with simulation predictions and
engineering limits to test whether the repaired hardware performs as predicted.
Discrepancies will guide refinements to the simulation models and checks.

\emph{(ii)~Transfer to another converter topology.} We will extend PowerFix
to boost or LLC by building its PowerDiagGraph, registering repair
operations, and checking the editable-field authorization map. We will
measure repair success and simulation cost on failed designs and report
adaptation effort in person-days, identifying which graph nodes and rules
were reused or newly authored.

\paragraph{Deployment risk.}
A patch can pass every encoded check yet miss a requirement,
such as supplier lead time. Authorization limits where patches can write,
and rollback can undo them. Neither guarantees that the encoded checks
are complete.
Engineer review and hardware checks remain necessary before release.

\section{Limitations}
\label{sec:limitations}

Results cover flyback designs and the encoded checks; the twelve natural
failures come from one generator. We cannot fully verify model access
across methods. Records confirm a shared endpoint and requested alias
for 24 of 96 primary tasks; the other 72 are matched only at the
documented model-family level (Appendix~\ref{app:model_provenance}).
On those 24 tasks, PowerFix verifies 17 designs and the Tool Agent five.
The separate backbone study spans 1.9 VRSR points across twelve models.
Engineer review time and hardware performance remain unmeasured.

\section{Conclusion}
\label{sec:conclusion}
PowerFix treats repair as more than resolving the reported fault: every required engineering check must also pass. It combines physical diagnosis, authorized edits, and validation before acceptance, reaching 78.7\% group-weighted VRSR against 30.1\% for the strongest baseline. 

The ablation distinguishes finding repairs from rejecting invalid outputs. Removing diagnosis costs 53.2 VRSR points; removing isolated validation costs only 3.7 but hands back 22 invalid designs out of 96. Success rates alone do not show which invalid outputs were withheld. 

All accepted patches stay within 29 authorized fields and include rollback. Their records show engineers the proposed edits and supporting checks, while leaving sign-off to them. 

The evidence is offline and limited to flyback. The deployment plan will test whether simulated repairs translate to hardware and other converter designs.

\clearpage
\bibliography{references}

\clearpage
\appendix
\raggedbottom
\setcounter{secnumdepth}{1}
\renewcommand{\thesection}{\Alph{section}}
\setcounter{table}{0}
\setcounter{figure}{0}
\renewcommand{\thetable}{S\arabic{table}}
\renewcommand{\thefigure}{S\arabic{figure}}
\section{Benchmark Construction}
\label{app:benchmark}

Table~\ref{tab:benchmark-construction} defines the specification and fault
codes used in the task records. Each healthy input--output combination is
used once for each fault class. Within every class, four tasks are mild,
four moderate, and four severe. The injected fields and reference repairs are kept in private
annotations. Reported failures and their evidence remain available
to the methods.

\begin{table}[!htb]
\centering
\caption{Independent choices in the controlled benchmark. Any input can
be paired with any output and fault class.}
\label{tab:benchmark-construction}
\small
\begin{tabularx}{\columnwidth}{@{}lX@{}}
\toprule
Code & Specification or fault \\
\midrule
I1 & 85--132\,Vac \\
I2 & 180--265\,Vac \\
I3 & 85--265\,Vac \\
\midrule
O1 & 5\,V / 2\,A (10\,W) \\
O2 & 12\,V / 1\,A (12\,W) \\
O3 & 12\,V / 2\,A (24\,W) \\
O4 & 24\,V / 1\,A (24\,W) \\
\midrule
F1 & Transformer turns-ratio selection \\
F2 & Magnetizing-inductance selection \\
F3 & Magnetic-component realization \\
F4 & MOSFET selection \\
F5 & Secondary-rectifier selection \\
F6 & Output-capacitor selection \\
\bottomrule
\end{tabularx}
\end{table}

\paragraph{Construction checks.}
The retained construction reports record 16 PLECS executions for the
healthy bases and 1,512 for controlled-fault construction. The latter
rejected 1,309 candidate faults before freezing the 72 tasks. Rejection
reasons include a missing target violation, unrelated critical failures,
and failed or incomplete validation; a candidate can have several reasons.
The packaging audit checked all 72 public tasks and their evidence hashes
and found no private ground-truth files in the public copies. These are
construction costs, excluded from the repair costs in Table~\ref{tab:main}.

\paragraph{Coupled faults.}
Table~\ref{tab:coupled-tasks} lists the four interaction patterns. Each
pattern is tied to one output specification and instantiated at all three
input ranges. Thus the data do not separate the effect of fault-pair type
from that of output specification. Coupled and natural tasks are shuffled
together before assigning T0073--T0096; identifiers do not encode the group.

\begin{table}[!htb]
\centering
\caption{Coupled-fault patterns. Each row contributes three tasks,
one at each input range I1--I3.}
\label{tab:coupled-tasks}
\small
\begin{tabularx}{\columnwidth}{@{}llX@{}}
\toprule
Output & Faults & Interaction \\
\midrule
O1 & F5+F6 & Rectifier loss and output ripple \\
O2 & F1+F4 & Turns ratio and MOSFET voltage stress \\
O3 & F2+F4 & Peak current, MOSFET loss, and temperature \\
O4 & F1+F5 & Primary-device stress and rectifier reverse voltage \\
\bottomrule
\end{tabularx}
\end{table}

\paragraph{Natural-failure records.}
The source runs were completed on 22 July 2026 (UTC), with one retained
failed output for each input--output specification. The source-to-task
mapping and the original design and validation records are retained.
Preparing benchmark copies normalizes constraint labels and simulator
bindings; it does not inject faults. The selection configuration requires
an observed engineering failure and prevents access to private reference
repairs. These tasks were not used to develop or tune PowerFix.
The supplied records identify the selected designs, but do not enumerate
the full source pool and every exclusion. This is a held-out set of
log-derived failures, not a random sample of all generator failures.

\section{Baselines and Execution Limits}
\label{app:baselines}

Table~\ref{tab:method-capabilities} compares the repair mechanisms and
execution limits. These are not matched compute budgets.

\begin{table}[!htb]
\centering
\caption{Primary repair mechanisms and execution limits. All methods
undergo external scoring. PowerFix uses $R=0$: validation tries
preplanned candidates without another LLM planning call.}
\label{tab:method-capabilities}
\small
\begin{tabularx}{\columnwidth}{@{}l>{\raggedright\arraybackslash}X>{\raggedright\arraybackslash}X@{}}
\toprule
Method & Repair mechanism & Limit \\
\midrule
Open-loop & Regenerates the full design. & Three proposal calls and one integration call. \\
Self-Refine & Revises from validation feedback; an LLM parses the edits. & At most three refinement rounds. \\
Tool Agent & Calls engineering tools and edits fields directly. & At most 16 agent steps. \\
PowerFix & Uses PowerDiagGraph to limit repairs; a compiler writes the edits. & One plan and at most six candidate validations. \\
\bottomrule
\end{tabularx}
\end{table}

For primary single-fault repair, PowerFix can use component and magnetics
routes. Each route retains up to three candidates after deterministic
health filtering. The planner ranks all supplied routes. Coupled and
natural tasks instead use joint candidates, drawing from up to 48 options
per dimension. Validation stops at the first accepted candidate, so six
is an upper bound rather than the number executed on every task.

The ablation and backbone studies cap each dimension at eight options
before candidate construction and health filtering. The ablation selects
one or two permitted dimensions at $R=0$; the backbone study validates one
candidate per round at $R=5$. Their full-system references therefore
differ in route selection, filtering order, and available candidates, as
well as the option cap. Appendix~\ref{app:protocols} examines the observed
task differences. Appendix~\ref{app:budgets} gives the budget checkpoints.

\paragraph{Scoring and recorded randomness.}
The external scorer checks sealed waveform hashes and recomputes metrics
instead of trusting saved pass/fail flags. It starts PLECS when usable
records are absent. Static or analytical failures can stop scoring before
simulation. Its outputs are kept outside the repair loop.

The primary comparison retains one trajectory per method--task pair.
The base seed is 20260721; task-specific seeds are derived from benchmark
position and accompany the model requests. Task bootstrap intervals do
not measure repeated decoding on the same task. The resampling units are
specified in Appendix~\ref{app:protocols}; model-access records and prompt
contents are described in Appendices~\ref{app:model_provenance}
and~\ref{app:token_audit}.

\section{Full Problem Formalization}
\label{app:formalism}
We give the design state and acceptance rule behind Equation~\ref{eq:creq}.
Let $D_t$ be the converter design state at step $t$, containing its topology,
editable parameters, constraints, analytical results, and simulation waveforms.
Applying patch $P$ produces the candidate $D_P=D_t\oplus P$.

Let
\begin{equation}
\mathcal{I}_t = \{\iota_1, \iota_2, \ldots, \iota_m\}
\end{equation}
denote the structured failure representations associated with $D_t$; each
$\iota$ records a failed metric, its operating condition, the observed
value, the required target, and supporting evidence.

For converter topology $T$, the topology-specific repair environment is
\begin{equation}
\mathcal{E}_T = \left(\mathcal{G}_T, \Omega_T, \mathcal{B}, \mathcal{V}_T\right),
\end{equation}
where $\mathcal{G}_T$ is PowerDiagGraph, $\Omega_T$ the permitted
repair operations, $\mathcal{B}$ the component database shared across
topologies, and $\mathcal{V}_T$ the executable analytical and simulation
tools. PowerDiagGraph is defined as
\begin{equation}
\mathcal{G}_T = \left(\mathcal{N}_T, \mathcal{A}_T\right),
\end{equation}
with $\mathcal{N}_T$ containing failed metrics, intermediate physical
quantities, and candidate causes, and $\mathcal{A}_T$ containing
decomposition and causal relations such as \emph{decomposed into},
\emph{determined by}, and \emph{caused by}.

An atomic patch is
\begin{equation}
P_t = \left(o_t, a_t, \Delta_t\right),
\end{equation}
with $o_t \in \Omega_T$ the selected operation, $a_t$ its arguments, and
$\Delta_t$ the resulting field-level modifications. Validating the
candidate produces
\begin{equation}
R_P = \operatorname{Validate}\!\left(D_P, \mathcal{C}_{\mathrm{req}}(P,\mathcal{I}_t)\right),
\end{equation}
and acceptance is
\begin{equation}
\operatorname{Accept}(P) = \operatorname{Valid}(P) \land \operatorname{ValidationPassed}(R_P),
\end{equation}
where $\operatorname{Valid}(P)$ requires the prescribed format and
preconditions and edits confined to authorized fields, and
$\operatorname{ValidationPassed}(R_P)$ requires every check in
$\mathcal{C}_{\mathrm{req}}$ to have been executed and to satisfy its
acceptance criterion. The state update is
\begin{equation}
D_{t+1} =
\begin{cases}
D_t \oplus P, & \text{if } \operatorname{Accept}(P),\\[3pt]
D_t, & \text{otherwise.}
\end{cases}
\end{equation}

\section{Method Details}
\label{app:method}

\paragraph{PowerDiagGraph composition.}
The flyback PowerDiagGraph contains 174 nodes and 301 directed arcs:
32 failed-metric, 76 mechanism, 14 observed-evidence, 25 design-field,
10 diagnostic-test, 10 validation-check, and 7 repair-operation nodes.
It is authored once per topology and fixed during inference.
The seven operation nodes describe topology-level repairs; they are not
an enumeration of the four experimental atomic operators, the four selection dimensions in the backbone study, or the nine Top-2 dimensions. These representations
have different roles and granularity. The four atomic operators replace
a MOSFET, rectifier, or output-capacitor bank, or redesign the magnetics.
Likewise, the 25 graph field nodes describe diagnostic relationships,
not the 29 editable paths in the experimental authorization map.
Graph check nodes support diagnosis; the runtime map determines which
checks execute.

\paragraph{A worked diagnosis.}
Table~\ref{tab:method-vds-evidence} gives the numerical evidence used
by the diagnostic test fixture.
This is an illustrative calculation, not an additional benchmark task.
The initial values support insufficient voltage margin and excessive
turn-off overshoot, while the plateau residual contradicts excessive
reflected voltage. Leakage and clamp absorption remain unresolved until
the additional measurements are supplied. With $L_{\rm lk}=200\,\mu$H
and turn-off current $10$ A, leakage energy is $10$ mJ. Overshoot also
rises by $27.4$ V per ampere between the two load measurements, above
the rule's $10$ V/A threshold. Together these rules identify excessive
leakage energy as the primary cause and also support insufficient clamp
absorption. The graph requests a registered
leakage/clamp discrimination experiment when these values are missing;
the fixture supplies them to test the rule outcomes.

\begin{table}[!htb]
\centering
\caption{Numerical MOSFET example. Thresholds belong to this fixture;
they are not universal device limits.}
\label{tab:method-vds-evidence}
\small
\begin{tabularx}{\columnwidth}{@{}Xrr@{}}
\toprule
Quantity & Value & Required \\
\midrule
Rating minus peak & $650-742=-92$ V & $\geq65$ V \\
Plateau residual & $|505-495|=10$ V & $\leq25$ V \\
Turn-off overshoot & $742-505=237$ V & $\leq120$ V \\
Leakage energy & $10$ mJ & $\leq4$ mJ \\
Clamp absorption fraction & $4/10=0.40$ & $\geq0.80$ \\
\bottomrule
\end{tabularx}
\end{table}

\paragraph{The field-to-check map.}
The experimental map has 58 associations over 29 paths. Each path maps
to both target-issue resolution and the full engineering regression,
so every candidate faces the same eight-operating-point check set.
The patch-dependent definition in Equation~\ref{eq:creq} also permits
finer selection. A separate map contains 43 checks and 368 associations
over 26 paths, but uses a different field schema and coverage: its paths
are not the experimental whitelist minus three entries. The schemas
and coverage would have to be reconciled before substitution. This
alternative is unused in the reported results, and no simulation saving
from selective checks is measured. The reported localization concerns
edits, not a reduced regression.

\paragraph{Interfaces and decision owners.}
Table~\ref{tab:method-interfaces} separates model decisions from the
programs that produce and check concrete edits.

\begin{table}[!htb]
\centering
\caption{Module interfaces. All rows except planning are deterministic;
PLECS supplies simulation evidence where needed.}
\label{tab:method-interfaces}
\small
\begin{tabularx}{\columnwidth}{@{}p{0.27\columnwidth}X@{}}
\toprule
Module & Input $\rightarrow$ output or decision \\
\midrule
Characterization & Design and reported failures $\rightarrow$ values, limits, conditions, and evidence links in a failure set. \\
Diagnosis & Failure set and fixed rules $\rightarrow$ supported, contradicted, or unresolved causes and available repairs. \\
LLM planning & Diagnosis and legal choices $\rightarrow$ route ordering or dimension selection. \\
Synthesis & Selected choices and frozen catalogs $\rightarrow$ concrete candidate intents. \\
Compilation & Intent and current state $\rightarrow$ exact field edits and their reversals. \\
Validation/gate & Patch, map, and design copy $\rightarrow$ check results and commit decision. \\
External scorer & Final output and separate task records $\rightarrow$ evaluation, never repair feedback. \\
\bottomrule
\end{tabularx}
\end{table}

\paragraph{Compilation preconditions.}
Before simulation, the compiler and operator catalog check the base-state
hash, operator registration and version, topology and design-family
scope, exact argument names, and numeric bounds. They require the
contract's component facts and any physical-applicability evidence.
Duplicate paths, missing or extra writes, and writes to protected or
derived fields are rejected. Every target must resolve to an existing
scalar field. Compilation records its old value, inverse edit, and
specification/topology hashes; the gate checks those invariants.

\paragraph{Recorded trace: T0053.}
The 24-V, 1-A task links its reproduced capacitor-margin failure to a
state hash and eight sealed waveform records. The public limit is
$10\times0.8=8$ V; the diagnostic rule uses the observed maximum
$24.28418$ V and obtains a margin of $-16.28418$ V. It identifies the
voltage-rating cause without an additional diagnostic experiment.
The LLM selects the component route. Three candidates are compiled;
two are validated before acceptance. The first fails the steady-state
drift check at both low-line, minimum-load bus corners: $0.66770$ and
$0.66530$ V, against $0.24$ V. The second passes the full regression.
Its analytical ripple estimate is $0.01646$ V; the maximum simulated,
linearly detrended ripple is $0.069509$ V, also below $0.24$ V.
Its explicit two-field write set includes the unchanged quantity
$7\!\rightarrow\!7$; only one field changes value. Rollback records
both fields, distinguishing the write footprint from the changed-value
count.

The formal run records 36 PLECS executions: 12 for failure reproduction
and 12 for each validated candidate. Each sweep covers eight corners;
four light-load corners require longer settling and a second execution.
The final external scorer checks hashes and recomputes metrics from
sealed waveforms, making zero fresh simulation calls in this case.
These counts use the completed run and its validation reports, not
three earlier rejected events retained in the same history.

\section{Detailed Evaluation and Accounting}
\label{app:protocols}

\paragraph{Fault classes.}
Table~\ref{tab:fault-family} separates the single-fault results by the
injected design error. Each column covers all twelve input/output
specifications, so differences are not caused by unequal class sizes.
The classes differ in repair difficulty: baseline coverage is especially
low for turns ratio and magnetizing inductance. These counts describe
this constructed set, not the frequency of faults in service.

\begin{table}[!htb]
\centering
\caption{Verified single-fault tasks by class, out of twelve per column.
Fault codes are defined in Table~\ref{tab:benchmark-construction}.}
\label{tab:fault-family}
\small
\begin{tabular*}{\columnwidth}{@{\extracolsep{\fill}}lrrrrrr@{}}
\toprule
Method & F1 & F2 & F3 & F4 & F5 & F6 \\
\midrule
Open-loop & 0 & 0 & 1 & 1 & 0 & 0 \\
Self-Refine & 3 & 3 & 4 & 4 & 12 & 7 \\
Tool Agent & 0 & 1 & 9 & 4 & 10 & 11 \\
PowerFix & 10 & 11 & 12 & 11 & 12 & 12 \\
\bottomrule
\end{tabular*}
\end{table}

\paragraph{Which methods repair the same tasks.}
Table~\ref{tab:repair-overlap} partitions all 96 tasks by the exact set
of methods that verify them. The intersections show both shared coverage
and repairs missed by PowerFix: Self-Refine alone verifies two single-fault
tasks, and Self-Refine and the Tool Agent verify one natural task that
PowerFix does not. The empty-set row contains tasks no method verifies.

\begin{table}[!htb]
\centering
\caption{Exact repair intersections. Only the listed methods verify a
task in each row; S, C, and N denote single, coupled, and natural tasks.}
\label{tab:repair-overlap}
\small
\setlength{\tabcolsep}{4pt}
\begin{tabularx}{\columnwidth}{@{}Xrrr@{}}
\toprule
Methods that verify & S & C & N \\
\midrule
PowerFix & 19 & 8 & 4 \\
Tool Agent, PowerFix & 17 & 1 & 2 \\
Self-Refine, Tool Agent, PowerFix & 18 & 1 & 0 \\
Self-Refine, PowerFix & 12 & 1 & 0 \\
None & 2 & 1 & 5 \\
Self-Refine & 2 & 0 & 0 \\
Open-loop, Self-Refine, PowerFix & 1 & 0 & 0 \\
Open-loop, PowerFix & 1 & 0 & 0 \\
Self-Refine, Tool Agent & 0 & 0 & 1 \\
\midrule
Total & 72 & 12 & 12 \\
\bottomrule
\end{tabularx}
\end{table}

\paragraph{New violations versus remaining violations.}
RIRR counts resolved targets whose final designs still fail other checks;
those failures need not be new. A stricter audit compares the initial and
final failed-check sets. If these are $F_{\rm initial}$ and
$F_{\rm final}$, a newly failed non-target check belongs to
\[
F_{\rm final}\setminus
  \bigl(F_{\rm initial}\cup\mathcal C_{\rm target}\bigr).
\]
The initial set combines observed failures with initial qualification
failures. For single-fault tasks, target exclusion also matches the
expected-check family; coupled and natural tasks use normalized target
identifiers. Check matching removes only terminal run suffixes such as
\texttt{:r1} and \texttt{:x1}. It retains the metric and its line,
load, and bus condition, so a failure at a new operating corner counts.
Table~\ref{tab:new-regressions} uses raw task counts rather than
group-weighted rates.

\begin{table}[!htb]
\centering
\caption{Tasks with at least one newly failed non-target check.
Denominators include only records with an evaluable final design.}
\label{tab:new-regressions}
\small
\begin{tabular*}{\columnwidth}{@{\extracolsep{\fill}}lrrrr@{}}
\toprule
Method & Single & Coupled & Natural & Total \\
\midrule
Open-loop & 47/72 & 9/12 & 8/12 & 64/96 \\
Self-Refine & 36/72 & 8/12 & 10/12 & 54/96 \\
Tool Agent & 21/72 & 6/12 & 5/12 & 32/96 \\
PowerFix & 0/68 & 1/12 & 2/12 & 3/92 \\
\bottomrule
\end{tabular*}
\end{table}

The three PowerFix cases are T0086, T0088, and T0096: rejected attempted
designs retained for external scoring, not accepted patches. Four
single-fault tasks (T0037, T0045, T0052, T0065) have no final design for
this audit and are marked unavailable, not zero. Their default
\texttt{frozen\_field\_violation} flags therefore do not demonstrate
four actual writes to protected fields. This audit is separate from
the validation-removal experiment in the main text.

\paragraph{Ablation reference and route differences.}
Table~\ref{tab:ablation_groups} gives the group counts behind the
ablation. The seven single-fault tasks repaired by the primary system
but not its ablation reference are all turns-ratio faults. On T0003,
T0028, T0048, T0055, T0064, and T0071, the primary system uses magnetics
repair; the ablation reference selects only rectifier replacement.
An offline reconstruction with the current generator finds no healthy
rectifier-only candidate even with all eleven rectifier options.

For T0033, the ablation reference selects magnetics and output capacitor.
The primary system's accepted design matches the tenth reconstructed
magnetics candidate; none of the first eight is healthy. This is a
current-code reconstruction, not a historical rank: the primary log
ranks that candidate first after health filtering. Route selection and
candidate truncation are thus distinct obstacles. These records do not
isolate an option-cap effect.

\begin{table}[!htb]
\centering
\caption{Verified task counts in the component ablation at $R=0$.
The main figure reports the group-weighted rates and paired differences.}
\label{tab:ablation_groups}
\small
\setlength{\tabcolsep}{3pt}
\begin{tabularx}{\columnwidth}{@{}Xrrr@{}}
\toprule
Configuration & \shortstack{Single\\/72} &
\shortstack{Coupled\\/12} & \shortstack{Natural\\/12} \\
\midrule
Full PowerFix & 61 & 10 & 6 \\
w/o Failure characterization & 31 & 3 & 4 \\
w/o Graph-guided diagnosis & 36 & 1 & 0 \\
w/o Constrained repair generation & 44 & 3 & 2 \\
w/o Isolated validation & 59 & 10 & 5 \\
\bottomrule
\end{tabularx}
\end{table}

\paragraph{Resampling implementation.}
The primary intervals use 100,000 bootstrap samples. Single and coupled
tasks share a draw of twelve input--output specifications; natural tasks
are resampled separately. The backbone intervals use 10,000 paired draws
of the twelve recorded base-design groups. Each group carries six single,
one coupled, and one natural task together across models. Both procedures
recompute the three group rates in each draw, preserving the pairing for
method differences.

\paragraph{Simulation counters.}
\label{sec:resource_accounting}
\emph{Workflow simulations} are the calls logged by each repair workflow,
used in the primary comparison and budget study. For PowerFix they
include failure reproduction and candidate validation, but exclude
external scoring. \emph{Logical operating-point evaluations} count
requests at individual line/load/bus conditions, including cache hits.
A sweep can contribute eight requests; settling extensions count
separately. The selection and backbone studies use this counter.
\emph{Fresh executions} count PLECS runs that actually execute and
exclude cache hits. These counters have different scopes and cannot be
substituted for one another. Fresh counts and active repair time depend
on the shared cache and execution order. The primary comparison lacks
complete active-time records across all methods and tasks.

The token audit is in Appendix~\ref{app:token_audit}; offline graph and
authorization-map authoring is outside these per-task counters.

\section{Candidate and Replanning Budgets}
\label{app:budgets}
\label{sec:results_scaling}

Tables~\ref{tab:candidate_budget} and \ref{tab:replanning-checkpoints}
give exact checkpoints behind the budget curves. $A$ limits candidate
validations within the initial plan; $R$ permits additional planning
rounds after validation feedback. A round may validate several candidates.
The $A=6$ and $R=0$ rows reproduce the primary configuration's
group counts.

\begin{table}[!htb]
\centering
\caption{Candidate limit under the initial plan ($R=0$). Group columns
give raw verified counts; VRSR is group-weighted.}
\label{tab:candidate_budget}
\small
\begin{tabular*}{\columnwidth}{@{\extracolsep{\fill}}rrrrr@{}}
\toprule
$A$ & Single /72 & Coupled /12 & Natural /12 & VRSR (\%) \\
\midrule
1 & 61 & 8 & 4 & 61.6 \\
2 & 64 & 9 & 4 & 65.7 \\
3 & 66 & 10 & 5 & 72.2 \\
4 & 67 & 10 & 6 & 75.5 \\
6 & 68 & 11 & 6 & 78.7 \\
\bottomrule
\end{tabular*}
\end{table}

\begin{table}[!htb]
\centering
\caption{Replanning checkpoints. S/C/N are verified counts out of
72/12/12. Costs are group-weighted means per task, including failures;
LLM calls exclude provider retries as separate calls.}
\label{tab:replanning-checkpoints}
\small
\setlength{\tabcolsep}{3pt}
\begin{tabular*}{\columnwidth}{@{\extracolsep{\fill}}rlrrrr@{}}
\toprule
$R$ & S/C/N & \shortstack{VRSR\\(\%)} &
\shortstack{Workflow\\simulations} & \shortstack{Fresh\\executions} &
\shortstack{LLM\\calls} \\
\midrule
0 & 68/11/6 & 78.7 & 22.13 & 22.13 & 1.00 \\
1 & 69/11/7 & 81.9 & 26.94 & 26.17 & 1.21 \\
2 & 70/11/8 & 85.2 & 31.06 & 29.61 & 1.39 \\
3 & 71/11/9 & 88.4 & 33.92 & 31.97 & 1.54 \\
5 & 71/11/9 & 88.4 & 35.79 & 33.61 & 1.77 \\
\bottomrule
\end{tabular*}
\end{table}

The grids are reported checkpoints. Omitting $R=4$ does not mean the
runner skips that round. These curves report point estimates;
confidence intervals are not shown.

Group weighting gives one additional natural repair six times the
contribution of one additional single-fault repair. The unchanged
coupled count at later checkpoints does not establish an exhaustive
search or identify why each remaining task stops. These budget results
also do not provide an equal-resource comparison with the primary
baselines.

\section{Repair-Dimension Selection}
\label{app:top2}

\paragraph{Policy definitions.}
Each policy selects two distinct permitted dimensions per planning
round. The nine dimensions are turns ratio, magnetizing inductance,
magnetic realization, MOSFET, rectifier, output capacitor, RCD clamp,
input stage, and switching frequency. The fixed policy uses this priority:
MOSFET, output capacitor, rectifier, magnetic realization, RCD clamp,
turns ratio, magnetizing inductance, input stage, switching frequency.
After a failure it selects an untried permitted pair. The random policy
samples untried permitted pairs using the task-specific seed and round
index. GPT-5 Mini reads the diagnosis, permitted dimensions, validation
feedback, and previously failed pairs.

\paragraph{Task and seed counts.}
The 31 tasks admit multiple repair dimensions: seventeen single-fault,
six coupled-fault, and eight natural failures. Five seeds give each
policy 155 trajectories, with group denominators 85, 30, and 40.
Table~\ref{tab:planner} retains both raw counts and VRSR because they
weight coverage differently. At $R=1$, for example, the fixed policy
verifies more trajectories than random selection but has lower VRSR;
more of its successes lie in the largest group.

\begin{table}[!htb]
\centering
\caption{Verified trajectories, with group-weighted VRSR (\%) in
parentheses. Each policy has 155 trajectories.}
\label{tab:planner}
\small
\setlength{\tabcolsep}{2pt}
\begin{tabular*}{\columnwidth}{@{\extracolsep{\fill}}lrrrr@{}}
\toprule
Policy & $R=0$ & $R=1$ & $R=2$ & $R=3$ \\
\midrule
Fixed & 55 (27.4) & 80 (39.4) & 95 (54.7) & 100 (60.2) \\
Random & 50 (26.9) & 78 (44.0) & 93 (53.9) & 118 (69.2) \\
GPT-5 Mini & 91 (53.7) & 125 (77.4) & 133 (81.7) & 136 (84.7) \\
\bottomrule
\end{tabular*}
\end{table}

\begin{table}[!htb]
\centering
\caption{Group-weighted resource use per trajectory at $R=3$,
including failures. Simulation counters follow Appendix~\ref{app:protocols}.}
\label{tab:planner-costs}
\small
\begin{tabular*}{\columnwidth}{@{\extracolsep{\fill}}lrrr@{}}
\toprule
Policy & \shortstack{Logical\\evaluations} &
\shortstack{Fresh\\executions} & \shortstack{Active\\time (s)} \\
\midrule
Fixed & 61.32 & 55.02 & 124.6 \\
Random & 61.12 & 55.09 & 124.6 \\
GPT-5 Mini & 40.15 & 37.43 & 162.6 \\
\bottomrule
\end{tabular*}
\end{table}

The policy curves report point estimates; no significance tests
are reported.
Active time includes planning and depends on the execution conditions.

\paragraph{Separate full-system reference.}
Table~\ref{tab:selection-reference} records full PowerFix on the same
subset, with one trajectory per task. Group counts are 15/5/6 at $R=0$
and 16/5/7 thereafter, out of 17/6/8. This reference allows the full
admissible repair space, while Top-2 restricts selection to two dimensions.
The one- versus five-seed protocols make the comparison descriptive;
isolating the effect of that restriction requires matched seeds and
candidate budgets.

\begin{table}[!htb]
\centering
\caption{Full-system reference on 31 tasks, without seed replication.
Costs are group-weighted means per task.}
\label{tab:selection-reference}
\small
\setlength{\tabcolsep}{3pt}
\begin{tabular*}{\columnwidth}{@{\extracolsep{\fill}}rrrrrr@{}}
\toprule
$R$ & \shortstack{VRSR\\(\%)} & \shortstack{Tokens\\(k)} &
\shortstack{Logical\\evaluations} & \shortstack{Fresh\\executions} &
\shortstack{Active\\time (s)} \\
\midrule
0 & 82.2 & 2.74 & 22.57 & 22.57 & 95.2 \\
1 & 88.3 & 3.22 & 26.49 & 25.87 & 110.5 \\
2 & 88.3 & 3.56 & 29.00 & 28.00 & 120.6 \\
3 & 88.3 & 3.79 & 30.87 & 29.54 & 127.6 \\
\bottomrule
\end{tabular*}
\end{table}

\begin{samepage}
\section{Backbone Robustness}
\label{app:backbones}
\label{sec:results_backbone}

\begin{table*}[!t]
\centering
\caption{Outcomes and costs in the backbone study (first replicate). VRSR@1 is group-weighted
success after the first candidate; Single@6 gives raw verified counts at
the six-candidate limit. All models verify 10/12 coupled and 7/12 natural
tasks at that limit. Resource columns are group-weighted means per task
at the maximum budget, including failed tasks.}
\label{tab:backbone_groups}
\label{tab:backbone-costs}
\small
\setlength{\tabcolsep}{4pt}
\begin{tabular*}{\textwidth}{@{\extracolsep{\fill}}lrrrrrrr@{}}
\toprule
Model & \shortstack{VRSR@1\\(\%)} & \shortstack{Single\\@6} &
\shortstack{Logical OP\\evaluations} & \shortstack{Fresh\\executions} &
\shortstack{Tokens\\(k)} & \shortstack{API\\attempts} &
\shortstack{Active time\\(s)} \\
\midrule
ERNIE 4.5 Turbo & 68.5 & 63/72 & 18.05 & 3.92 & 18.55 & 2.38 & 104.3 \\
GPT-5.4 Nano & 68.5 & 63/72 & 18.31 & 2.58 & 15.15 & 2.35 & 79.1 \\
Qwen 3.7 Max & 68.5 & 62/72 & 18.78 & 3.06 & 19.16 & 2.36 & 125.3 \\
DeepSeek V4 Pro & 69.0 & 62/72 & 18.14 & 3.84 & 16.43 & 2.44 & 115.7 \\
Grok 4.1 & 68.5 & 62/72 & 18.18 & 3.81 & 15.09 & 2.55 & 157.8 \\
GPT-5 Mini & 68.1 & 62/72 & 18.70 & 4.21 & 17.46 & 2.40 & 148.5 \\
Qwen 3.5 Flash & 68.5 & 61/72 & 18.72 & 3.75 & 25.82 & 2.52 & 170.3 \\
Doubao Seed 1.6 & 68.5 & 61/72 & 18.31 & 3.69 & 19.79 & 2.36 & 140.9 \\
GLM-5.2 & 69.0 & 61/72 & 18.76 & 2.99 & 20.13 & 2.57 & 185.0 \\
Claude Sonnet 5 & 65.7 & 61/72 & 18.25 & 3.58 & 23.88 & 2.41 & 144.7 \\
Claude Opus 5 & 63.0 & 61/72 & 19.03 & 6.00 & 30.25 & 2.92 & 273.8 \\
Gemini 3.6 Flash & 68.5 & 59/72 & 18.76 & 3.86 & 20.66 & 2.40 & 108.6 \\
\bottomrule
\end{tabular*}
\end{table*}

The first replicate contains 1,152 trajectories: 96 tasks for each of
twelve models. An initial plan and up to five replanning rounds allow at most
six candidate validations, one per round, with up to eight options per repair
dimension. Search stops when a candidate is accepted. Table~\ref{tab:backbone_groups}
reports initial success, final single-fault coverage, and recorded resource
use under this protocol. VRSR@1 uses only the first candidate. The final
GPT-5 Mini reference is 75.9\%, as in Figure~\ref{fig:backbone-main}.

\end{samepage}

\paragraph{Attempt budgets and paired comparisons.}
The recorded checkpoints allow at most one, two, three, four, or six
candidates. These are budget limits, not the number each task requires.
After acceptance, later checkpoints retain the accepted outcome without
running another repair. The table separates initial selection from the
coverage obtained after replanning.

Qwen 3.7 Max and DeepSeek V4 Pro match GPT-5 Mini's final binary outcome on
every task in the first replicate. Their paired bootstrap differences
are therefore zero in every resample. These zero-width intervals describe
agreement on this task set; they do not remove uncertainty about other tasks
or decoding runs. Grok 4.1 has the same group totals as GPT-5 Mini but differs
on two tasks, giving a nonzero interval. The eleven alternative-model
intervals shown in the main text are separate, unadjusted 95\% intervals,
not a simultaneous 95\% guarantee for all comparisons.

\paragraph{Resource accounting.}
Simulation counters follow Appendix~\ref{app:protocols}. Fresh counts
and active repair time depend on the shared cache and execution order;
they describe these runs rather than intrinsic model speed.

Tokens include recorded input and output usage. API attempts include
provider retries, whereas a workflow-level LLM request can contain several
such attempts. GPT-5 Mini averages 2.40 API attempts and 2.34 workflow-level
requests per task. Appendix~\ref{app:token_audit} gives the prompt and token
accounting. No complete official billing export is available, so the token
column is not a comparison of monetary cost.

\paragraph{Replicate stability.}
The first replicate uses seed 20260721. A second replicate, with seed
20260722, covers a fixed subset of 24 tasks: twelve single-fault, six coupled,
and six natural tasks. This adds 288 trajectories, for 1,440 in total.
Every model has identical final binary outcomes across the two replicates
on this subset. First-acceptance indices also agree on all 24 tasks for ten
models, and on 23 of 24 for GPT-5 Mini and Grok 4.1; joint nonacceptance
counts as agreement. This checks repeatability on the selected subset,
not on all 96 tasks. It does not establish stability under provider updates
or identify the model snapshots missing from historical records.

\section{Model-Access Provenance}
\label{app:model_provenance}

The audit distinguishes requested aliases from gateway-returned identifiers.
For the primary coupled and natural batch (24 tasks across four methods),
runtime fingerprints confirm a shared endpoint,
\url{https://api.vveai.com/v1}, and requested alias
\texttt{gpt-5-mini}. Returned identifiers include both
\texttt{gpt-5-mini} and \texttt{gpt-5-mini-2025-08-07}; these strings
alone do not establish different upstream snapshots.

The backbone study uses the same gateway with model-specific aliases;
its GPT-5 Mini condition requests \texttt{gpt-5-mini-2025-08-07}.
Historical probes, condition fingerprints, and request metadata support
that routing provenance. Endpoint records are missing for the
single-fault batch and the earlier ablation and budget runs, which
predate the logging configuration. Those comparisons are matched at
the documented model-family level. Establishing a shared upstream
snapshot across batches would require provider records we do not hold.

\section{Token Usage and Request Audit}
\label{app:budget_audit}
\label{app:token_audit}

The audit covers 480 task results: four primary methods on 96 tasks
and the 96-task GPT-5 Mini backbone condition, with 2,084 workflow-level
call records. All result files match their frozen manifest or table
hashes. This confirms record identity, not upstream model identity
(Appendix~\ref{app:model_provenance}). Appendix~\ref{app:budgets}
gives the budget checkpoints.

\paragraph{Counting usage.}
Table~\ref{tab:token_breakdown} includes unsuccessful tasks. Tokens per
request divide group-weighted per-task tokens by requests, before
rounding. Usage may include retries within a request; missing usage is
not estimated. These online costs exclude graph and map authoring and
deterministic enumeration. They are not total ownership costs or an
equal-budget comparison.

\begin{table}[!htb]
\centering
\caption{Recorded token usage. Req./API gives workflow requests and
API attempts per task; API attempts include retries. Input and output
are tokens per request; total is thousands of tokens per task.
PowerFix conditions follow Table~\ref{tab:protocols-main}.}
\label{tab:token_breakdown}
\small
\setlength{\tabcolsep}{2pt}
\begin{tabular*}{\columnwidth}{@{\extracolsep{\fill}}lrrrr@{}}
\toprule
Condition & Req./API & Input & Output & Total \\
\midrule
\shortstack[l]{PowerFix\\(primary)} & 1.00/1.06 & 1,895.96 & 835.97 & 2.73 \\
\shortstack[l]{PowerFix\\(backbone)} & 2.34/2.40 & 6,215.58 & 1,238.18 & 17.46 \\
Open-loop & 4.00/4.44 & 6,535.81 & 2,738.94 & 37.10 \\
Self-Refine & 5.23/5.29 & 5,134.95 & 1,257.65 & 33.44 \\
Tool Agent & 11.08/11.32 & 6,865.63 & 613.14 & 82.86 \\
\bottomrule
\end{tabular*}
\end{table}

\paragraph{What the prompts contain.}
The primary planner receives check identifiers, numerical diagnosis,
evidence, and registered repair routes. It orders routes without
further LLM calls at $R=0$.

Open-loop roles repeatedly receive requirements, the failed design,
and failure information, with catalogs or earlier proposals where
relevant. Self-Refine analysis reads the full design, catalogs, and
feedback; edit parsing receives editable paths. The Tool Agent reads
the design, editable fields, tools, remaining budget, and last eight
tool-history entries.

The backbone planner reads the full design, failures, diagnosis, and
four dimensions: MOSFET, rectifier, output capacitor, and magnetics.
It selects one or two without expanding candidate lists in the prompt.
Later rounds add validation feedback and the last three attempts.
This differs from primary route ordering and nine-dimension selection.
Initial backbone planning averages 6,048.97 input and 1,023.13 output
tokens; later calls average 6,339.68 and 1,398.34. Later calls occur only
on tasks that have not succeeded, so this difference cannot isolate
feedback overhead.

\section{Accepted Patch Profile}
\label{app:patch-profile}

We profile all 85 accepted task submissions in the primary $R=0$
configuration: 68 single-fault, 11 coupled-fault, and 6 natural-failure
tasks. Each joint repair counts as one submission; only the final
selected route is included. Rejected candidates are not part of the
field-count distribution. Percentages use these 85 submissions as
the denominator, without equal-group weighting.

\paragraph{Counting fields.}
For each submission, we count distinct field paths explicitly present in
\texttt{primary\_edits}, taking their union across constituent patches.
A write is counted even when its recorded old and new values agree.
The measure therefore describes the recorded write footprint, rather
than the number of values that differ between the initial and final
design. Rollback entries, read-only attributes, and downstream
recomputed fields absent from the explicit edits are excluded.
Table~\ref{tab:patch-profile} gives every nonzero frequency; the JSON
also retains the zero-frequency bins. Quantiles use linear interpolation
at index $(n-1)q$, with $n=85$ and $q=0.9$ for P90. A threshold share
is the sum of counts at or below that threshold divided by 85.

\begin{table}[!ht]
\centering
\caption{Distribution of explicit field writes in accepted $R=0$
submissions. Each field path counts once per submission, including writes
that retain its current value.}
\label{tab:patch-profile}
\small
\begin{tabular*}{\columnwidth}{@{\extracolsep{\fill}}lrrrrr@{}}
\toprule
Explicit fields & 1 & 2 & 3 & 25 & 26 \\
Submissions & 28 & 13 & 3 & 34 & 7 \\
\bottomrule
\end{tabular*}
\end{table}

\paragraph{Write composition.}
The \emph{swap} class contains only writes to the MOSFET, rectifier,
output-capacitor, or magnetic-core part-number fields: 28 submissions
(32.9\%). The \emph{parameter} class contains no such reference writes:
zero submissions. The \emph{mixed} class contains both reference and
other field writes: 57 submissions (67.1\%). These labels describe
the explicit patch contents, not independently verified physical
replacements. Composition is assigned after taking the union of writes
in a joint submission, rather than separately for each atomic patch.

\paragraph{Operation coverage.}
The experimental catalog registers four atomic operators
(Table~\ref{tab:patch-operators}). These are distinct from the seven
topology-level repair-operation nodes in PowerDiagGraph
(Appendix~\ref{app:method}). An operator is counted once per submission
in which it appears. Ten submissions combine two operators, so the
frequencies sum to 95 rather than 85.

\begin{table}[!ht]
\centering
\caption{Atomic operators in accepted $R=0$ submissions. Joint repairs
contribute to more than one row.}
\label{tab:patch-operators}
\small
\begin{tabular}{@{}lr@{}}
\toprule
Experimental operator & Submissions \\
\midrule
\texttt{swap\_mosfet} & 20 \\
\texttt{swap\_rectifier} & 18 \\
\texttt{swap\_output\_capacitor} & 16 \\
\texttt{redesign\_magnetics} & 41 \\
\bottomrule
\end{tabular}
\end{table}

\paragraph{Validation and rollback.}
No explicit field path falls outside the experimental authorization map,
and all 85 accepted submissions (100\%) contain rollback edits. This
checks the presence of rollback records, not a separate rollback
execution. Before the
selected repair was accepted, 83 submissions had no validation-rejected
candidate and two had one. This count includes earlier routes but
excludes compilation failures, preselection failures, and checks of
alternative routes after acceptance. Repeated historical trace events
are excluded using the executed-candidate counter and current-run
timestamps. It is consistent with $R=0$,
which permits candidate validation within the initial plan but no
feedback-driven replanning. This profile covers $R=0$ only.

\end{document}